\documentclass[conference]{IEEEtran}
\IEEEoverridecommandlockouts

\usepackage{cite}
\usepackage{amsmath,amssymb,amsfonts}
\usepackage{algorithmic}
\usepackage{graphicx}
\usepackage{textcomp}
\usepackage{xcolor}
\usepackage{booktabs}
\usepackage[hyphens]{url}
\usepackage{tikz}

\newcommand{\whitecircle}[1]{%
  \tikz[baseline=(char.base)]{
    \node[shape=circle,draw,inner sep=1pt,fill=white] (char) {#1};
  }
}

\def\BibTeX{{\rm B\kern-.05em{\sc i\kern-.025em b}\kern-.08em
    T\kern-.1667em\lower.7ex\hbox{E}\kern-.125emX}}
\begin{document}

\pdfpagewidth=8.5in
\pdfpageheight=11in

\newcommand{\iscasubmissionnumber}{107}

\pagenumbering{arabic}

\title{Mira: Memory-Efficient \textbf{M}oE \textbf{I}nfe\textbf{r}ence Using \textbf{A}daptive Caching and Predictive Expert Staging 
}


\author{
\IEEEauthorblockN{
Sanjali Yadav,
Bahar Asgari}
\IEEEauthorblockA{University of Maryland, College Park \\
\{sanjali7, bahar\}@umd.edu}
}

\maketitle
\thispagestyle{plain}
\pagestyle{plain}


\begin{abstract}

Mixture-of-Experts (MoE) models are a compelling architecture for scaling model capacity, making them especially attractive for deployment on resource-constrained, single-GPU systems. However, this benefit is difficult to realize because expert parameters dominate memory, and token-level routing is dynamic, unpredictable, and skewed. Prior work using offloading and caching remains fundamentally reactive as systems wait for router outputs before moving experts, leading to inefficient cache utilization, and an inability to overlap transfers with compute under tight VRAM budgets.

To address these challenges, we propose Mira, an algorithm-system co-design that enables high-capacity MoE inference on a single GPU. Mira shifts from a reactive to a proactive stance by coupling predictive expert management with a tailored quantization format. It introduces lightweight per-layer predictors that anticipate expert usage two layers ahead, enabling proactive prefetching. These predictions feed a two-tier HOT+STAGE GPU cache managed by token-level routing telemetry to retain frequently used experts while staging predicted ones. To minimize transfer overhead, Mira implements a custom compression for expert parameters, which reduces metadata and improves packing efficiency, while minimally degrading accuracy.

Mira is implemented as a fully integrated runtime that coordinates predictors, caching policies, and quantized transfers to maximize overlap between communication and compute. Our experiments show that Mira reduces expert-induced stalls. Compared against state-of-the-art baselines, Mira achieves a 5.71$\times$ speedup in average throughput on a memory-constrained GPU. It accelerates Time-to-First-Token by 11.71$\times$  and achieves a 3.84$\times$ average speedup in beam search inference, demonstrating its effectiveness across diverse inference scenarios.

\end{abstract}

\section{Introduction}
\label{sec:intro}

Large Language Models (LLMs) deployed as chat-style interfaces~\cite{chatbot_survey} have become a standard part of modern software stacks, powering search, productivity tools, programming assistants, and consumer applications. The dominant deployment model involves large-scale distributed inference systems, which are the focus of substantial research and capital investment aimed at serving millions of users \cite{patel2023inference,wu2023fast}. Concurrently, a compelling need for local, on-device LLM inference is emerging \cite{localai,song2024powerinfer,gpt4all}. This trend is driven by critical requirements that cloud-centric systems cannot easily satisfy: data privacy and security, particularly when handling sensitive healthcare or government data \cite{Zylon_PrivateGPT_2023}; the ability to personalize models through fine-tuning on confidential datasets \cite{lyu2024llm}; and democratizing access to LLMs, especially for users without access to high-end GPUs or large-scale compute clusters.

However, executing local inference presents a significant system optimization challenge driven by the vast disparity between the memory footprint of the state-of-the-art models, which often contain billions of parameters, and the limited VRAM in typical consumer-grade single-GPU setups. The local inference environments lack the terabytes of high-bandwidth memory and massive parallelism found in industry-scale accelerator clusters, making the task to optimize them non-trivial, as every available gigabyte of VRAM must be optimally utilized to accommodate the model's parameters and activation cache.  

Additionally, the quest for higher generation quality has driven a massive increase in model scale, which in turn has escalated computational requirements. To address these high compute demands, Mixture of Experts (MoE) models have been proposed as an alternative to traditional dense LLMs \cite{du2022glamefficientscalinglanguage,fedus2022switchtransformersscalingtrillion,shazeer2017outrageouslylargeneuralnetworks}. The primary advantage of the MoE architecture is its ability to decouple model capacity from computational cost; unlike dense models that activate all parameters for every token, MoE models employ sparse activation where a trained router network contextually selects a small subset of experts to process each token. This allows the model to scale its parameter count significantly while keeping the per-token compute cost manageable. 

However, the strengths of MoE at the algorithm level translate into acute system-level pressure on memory and I/O. The aggregate parameter count of the experts dominates the model’s memory footprint, and the token-dependent, sparse activation pattern causes highly skewed and dynamic expert usage. On resource-constrained, single-GPU setups, this combination creates a challenge: experts cannot all reside on GPU in high precision, yet loading them just-in-time from CPU DRAM risks adding PCIe latency on the critical path.

Prior work largely addresses this memory challenge through two orthogonal techniques: expert offloading \cite{du2024sida,hwang2024pre,kamahori2024fiddler,shazeer2017outrageouslylargeneuralnetworks} and quantization \cite{frantar2023qmoe,kim2023mixturequantizedexpertsmoqe,kim2022says}. Offloading keeps expert parameters in CPU memory and migrates only the selected experts to GPU when the router activates them. Quantization reduces the bit-precision of parameters to shrink their storage footprint and accelerate transfers over the PCIe bus. Recent systems combine both ideas \cite{eliseev2023fast,zhou2025floe,rajbhandari2022deepspeed,tang2024hobbit}, aggressively quantizing experts while offloading them to CPU, and then using general-purpose caching or eviction policies to decide which experts to keep resident on GPU.

Yet, this hybrid approach leaves an optimization gap. First, existing systems typically treat quantization as a passive compression step and ignore the rich, token-level routing signals available in MoE, resorting to coarse-grained policies such as recency-based eviction \cite{xue2025moeinfinityefficientmoeinference,sarkar2023edge,eliseev2023fast}, or static popularity heuristics \cite{kamahori2024fiddler,xue2025moeinfinityefficientmoeinference,yao2024exploiting}. Second, expert offloading is often reactive: the system learns which experts are needed only after the gate has already made its decision, forcing the runtime into an unfavorable trade-off. Fetching experts late incurs stalls on CPU to GPU transfers, while speculative prefetching without high accuracy wastes VRAM on experts that may never be used. The alternative relies on ultra-low-bit quantization to keep all experts on GPU, trading I/O stalls for potential degradation in model quality.

To address these limitations, we introduce \textbf{Mira}\footnote{\textbf{Mira} is a blue subgiant star in the constellation Musca}, an algorithm-system co-design for MoE inference on memory-constrained GPUs. Mira couples modest, accuracy-preserving post-training quantization with an adaptive, prediction-driven expert cache that is explicitly aware of token-level routing behavior. At the algorithm level, Mira trains lightweight per-layer predictors that learn to anticipate which experts will be activated in future MoE layers. At the system level, Mira uses these predictions to pre-stage just the right experts into GPU memory, managing VRAM as a two-tier HOT+STAGE cache backed by a CPU-resident, INT8-compressed expert store. 

Concretely, Mira makes three key design choices:

\begin{itemize} 

\item \textbf{Quantization-aware expert store:} Mira applies a custom INT8 post-training quantization exclusively to MoE expert parameters, while keeping non-expert layers in higher precision. All experts are stored as CPU-pinned compressed tensors, enabling high-throughput CPU to GPU transfers and a unified representation across offloaded, staged, and hot-resident experts. This design avoids ultra-low-bit quantization that risks quality loss, while still reducing the memory and bandwidth cost per expert. 

\item \textbf{Predictive expert staging:} Mira augments each MoE layer with a small predictor that uses features derived from the current hidden states, router logits, and recent routing history to forecast the top experts required in the next MoE layers (e.g., an $i \rightarrow i\!+\!2$ pattern). These predictions drive a speculative prefetcher that stages only the most likely experts into GPU memory, overlapping CPU to GPU transfers with ongoing computation and reducing PCIe-induced stalls. 

\item \textbf{Adaptive HOT+STAGE caching under VRAM budgets:} Mira partitions GPU memory into a stable HOT region for long-lived, frequently used experts and a smaller STAGE window for predicted experts. At execution time, Mira maintains lightweight per-expert metrics (e.g., routed volume and cache hit/miss behavior) and uses them to continuously rebalance the HOT set. This metric-driven accounting allows Mira to evict underutilized experts from GPU, promote frequently used experts from CPU to GPU, and adapt to changing workloads without manual tuning.  
\end{itemize}

By tightly integrating predictive expert selection with a quantization-aware caching and staging runtime, Mira enables high-capacity MoE models to run efficiently on a single, memory-constrained GPU. Our evaluation shows that Mira reduces expert-migration stalls and improves end-to-end throughput over prior offloading schemes, while maintaining model quality and supporting larger MoE configurations within the same VRAM budget. Our evaluation shows that Mira improves end-to-end throughput by 10.46$\times$ over DeepSpeed and 2.86$\times$ over MoE-Infinity on a 48GB GPU, and by 5.71$\times$ over MoE-Infinity on a 24GB GPU, while reducing time-to-first-token by up to 18.38$\times$ and achieving up to a 3.84$\times$ speedup over Fiddler in beam-search decoding, all with accuracy close to an FP16 baseline on downstream tasks.

\section{Background}
\label{sec:background}

\subsection{Mixture of Experts}
\label{sec:bg-moe}

Dense LLMs adopt a uniform computation pattern in which every token is processed by all layers and all parameters of the model. In a standard Transformer-based LLM \cite{vaswani2017attention}, each layer consists of self-attention, normalization and a feed-forward network (FFN); for each input token, all the components are executed regardless of the type of token. As a result, the per-token compute cost scales linearly with the number of parameters: increasing model capacity (e.g., wider FFNs or more layers) directly increases both training and inference FLOPs. This tight coupling between capacity and compute has driven interest in architectures that can increase the number of parameters without proportionally increasing the cost of serving each token.

\begin{figure}[h]
    \centering
    \includegraphics[width=0.5\textwidth, scale=1.2]{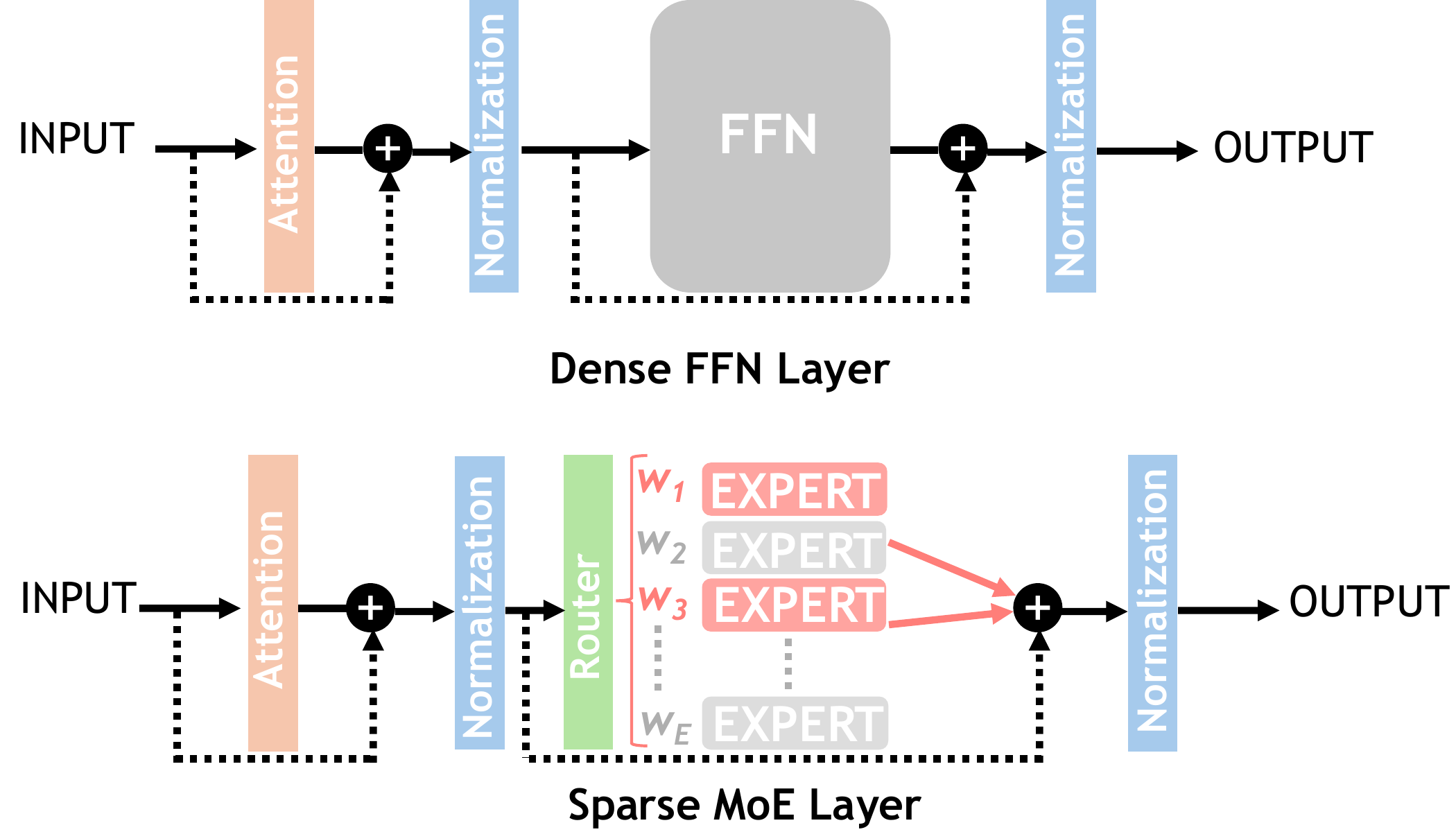}
    \vspace{-12pt}
    \caption{The architecture of dense transformer block compared to sparse MoE block.}
    \label{fig:moe_intro}
    \vspace{-10pt}
\end{figure}

MoE models provide such an architecture by introducing conditional computation in place of the dense FFN blocks. In a typical MoE layer, illustrated in Figure~\ref{fig:moe_intro}, the single FFN is replaced by a pool of $E$ experts, each an FFN with its own parameters. The router network computes a score $W$ over experts for each token and activates only the top-$k$ experts. In Figure~\ref{fig:moe_intro}, top-$k$ is set to 2. The token is then processed by the selected subset of experts, and their outputs are combined. Because only $k \ll E$ experts are active per token, the per-token compute and memory bandwidth cost can be kept close to that of a dense layer, while the total parameter count of the model grows roughly with $E$. This conditional execution decouples \emph{capacity} from \emph{per-token compute}. MoE models can achieve effective model sizes in the hundreds of billions or trillions of parameters while maintaining inference-time FLOPs comparable to much smaller dense models \cite{fedus2022switchtransformersscalingtrillion,du2022glamefficientscalinglanguage}. 

Prior work shows that this additional capacity can improve perplexity and downstream task performance at fixed or modestly increased compute budgets, and can shift the empirical scaling laws of LLMs in a favorable direction \cite{du2022glamefficientscalinglanguage, shazeer2017outrageouslylargeneuralnetworks}. From an algorithmic perspective, the experts can specialize on different regions of the input space, languages, or tasks, and the router learns to allocate computation accordingly. At the same time, MoE models introduce new challenges that do not arise in dense LLMs. First, while only a few experts are active per token, all experts must be stored somewhere in the system, and their aggregate parameter count dominates the memory footprint of the model. 

In large MoEs models, the experts often account for the majority of parameters in the network, far exceeding the attention and non-expert layers. In large-scale datacenter training, this motivates distributing experts across many devices and using all-to-all communication to route tokens to remote experts \cite{lepikhin2020gshard}. In single-node inference scenarios, it creates pressure on GPU memory capacity and PCIe bandwidth, since most experts are idle for any given token but must remain available for routing decisions. Second, the token-dependent routing behavior introduces load imbalance and dynamism that are absent in dense models. 

Dense LLMs have a fixed and predictable compute graph: every token incurs the same sequence of operations, enabling straightforward static scheduling and batching. In contrast, the MoE router produces a highly skewed and evolving distribution of expert usage, where some experts may be hot and frequently selected, while others are rarely used except on specific inputs or languages. This skew impacts both algorithmic considerations (e.g., load-balancing losses during training) and systems concerns (e.g., which experts should be kept resident on GPU during inference, and which can be offloaded to CPU or disk without incurring stalls).

\begin{figure*}[h]
\vspace{-0pt}
  \includegraphics[width=2.1\columnwidth, scale=0.05]{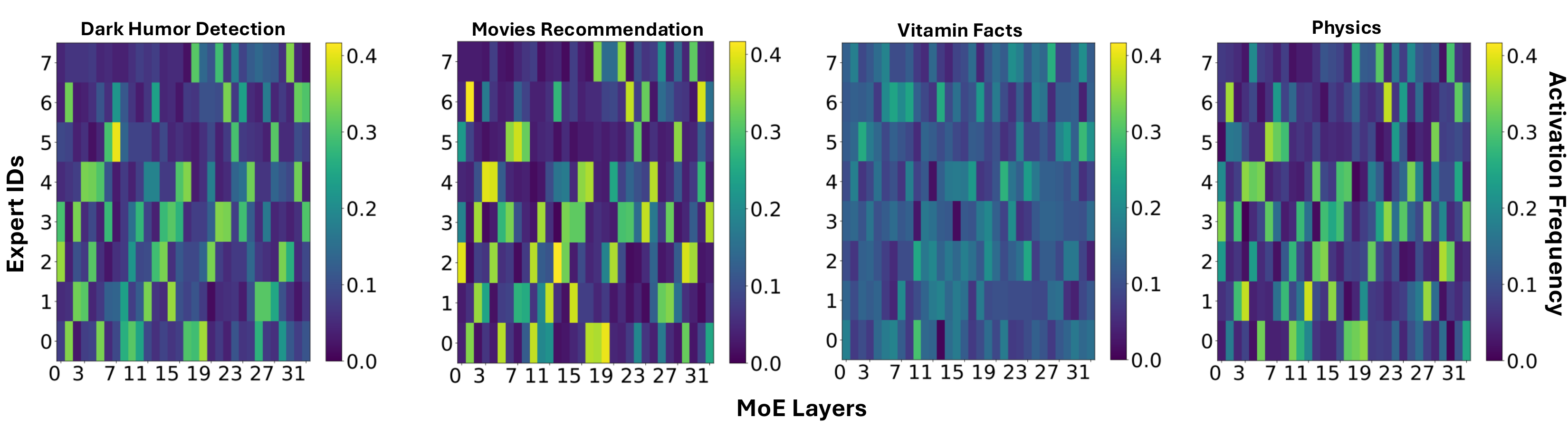}
  \vspace{-20pt}
  \caption{Aggregated expert activation patterns for Mixtral-8x7B ~\cite{hendrycks2020measuring} on four distinct MMLU tasks . Each heatmap shows the activation frequency (color) for each of the 8 experts (y-axis) across the 32 MoE layers (x-axis).}
  \label{fig:expert_activation}
  \vspace{-10pt}
\end{figure*}

\subsection{LLM Quantization}
\label{sec:bg-quant}

Quantization is a standard technique for reducing the memory and compute cost of neural network inference by representing weights and activations in low-precision integer formats (e.g., INT8, INT4) instead of high-precision floating point \cite{dettmers2022gpt3,xiao2023smoothquant,lin2024awq,dong2023packqvit,rakka2025mixed}. In most LLM deployments this is implemented as uniform affine quantization, where each tensor or channel is mapped to a small integer range using a calibrated scale factor. 

At inference time, the common design is weight-only quantization where weights are stored in low precision and dequantized on the fly into higher-precision accumulators, or consumed directly by integer GEMM kernels with dequantization fused into the epilogue \cite{zhou2025floe, tang2024hobbit}. This shrinks the footprint of model parameters and the bandwidth required to move them between CPU and GPU, which is particularly important for single-GPU setups where VRAM and PCIe bandwidth are the dominant bottlenecks. 

Formally, post-training quantization (PTQ) typically uses uniform affine quantization. Given a full-precision tensor $w \in \mathbb{R}^n$ and a bit-width $b$, the integer range is $[q_{\min}, q_{\max}] = [-2^{b-1}, 2^{b-1}-1]$ for signed quantization. PTQ chooses a scale $s > 0$ and zero-point $z \in \mathbb{Z}$, and maps each element as

\begin{equation}
q_i = \mathrm{clip}\big(\mathrm{round}(w_i / s) + z, q_{\min}, q_{\max}\big),
\qquad
\hat{w}_i = s \cdot (q_i - z),
\end{equation}

where $q_i$ is the stored integer and $\hat{w}_i$ is the dequantized approximation. 


PTQ methods typically use per-channel or block-wise granularity for weights to better track local statistics, while leaving especially sensitive components such as embeddings, layer norms, and final output layers in FP16/BF16. MoE architectures amplify the benefits of this heterogeneous treatment: expert MLPs dominate the parameter count and are accessed sparsely, making them attractive targets for aggressive compression, whereas the shared backbone and router networks are relatively small but latency-critical and thus often kept at higher precision to preserve routing behavior and overall quality \cite{fu2025eaquant}.

\section{Motivation}
\label{sec:motivation}

\subsection{Quantization Tradeoff}
\label{sec:mot-quantization}

Table \ref{tab:quantization_perplexity} illustrates the core challenge in quantizing MoE experts for offloaded inference by reporting the measured perplexity on a subset of WikiText-2 under different quantization schemes. As the bit-width of expert weights is reduced from FP16 to INT8 and then to INT4, the perplexity increases, degrading generation quality. This degradation, however, comes with the benefit of substantially lower CPU to GPU transfer time, almost a 2$\times$ reduction as quantization becomes more aggressive, so more aggressive quantization directly translates into faster expert migrations. In practice, though, ultra-low-bit formats can spend a larger fraction of time in dequantization and related bookkeeping, especially in local inference settings with single-batch decoding, where there is insufficient compute to hide the dequantization overhead.

\begin{table}[h]
    \centering
    \caption{Perplexity on WikiText-2 for varying quantization bit widths.}
    \label{tab:quantization_perplexity}
    \resizebox{\columnwidth}{!}{%
    \begin{tabular}{lcccccc}
        \toprule
        & \textbf{FP16} & \textbf{INT8} & \textbf{INT8 (Mira)} & \textbf{INT4} & \textbf{INT3} & \textbf{INT2} \\
        \midrule
        \textbf{Perplexity $\downarrow$}
        & 3.9324 & 3.9695 & 3.9554 & 4.1048 & 5.2653 & 1147.5236 \\
        \bottomrule
    \end{tabular}%
    }
\end{table}

Importantly, the transfer volume is not solely determined by the nominal bit-width of the weights. Modern low-bit schemes rely on increasingly fine-grained scaling (e.g., per-channel or per-group scales, sometimes with additional outlier paths) to recover accuracy at 4 bits and below. As a result, an expert migration over PCIe consists not only of the quantized weight matrix but also of its associated scales and metadata, which may be stored at higher precision. Under aggressive group-wise quantization, the number of scale values grows with the number of groups, thus this metadata can become a non-trivial fraction of the overall transfer volume and an additional source of memory traffic and kernel complexity during dequantization \cite{xie2025amove}.

Many recent systems sidestep this tradeoff by co-designing quantization schemes with specialized accelerator support. Integer-only methods such as I-BERT~\cite{kim2021bert} and I-LLM~\cite{hu2024llm} quantize the entire transformer computation graph to low-bit integers and execute all operators using integer arithmetic on hardware INT8/INT4 units, explicitly avoiding any dequantization back to floating point during inference. Likewise, hardware-oriented PTQ frameworks such as ZeroQuant \cite{yao2022zeroquant} and SmoothQuant \cite{xiao2023smoothquant} target W8A8 or mixed INT4/INT8 formats that are executed directly by low-precision GEMM kernels on tensor-core GPUs, while recent FP8 recipes for LLMs rely on native FP8 Tensor Cores in Hopper-class accelerators to run transformer matmuls in FP8 end-to-end rather than treating FP8 as a mere storage format \cite{nvidia2025fp8,acceleratingllama3fp8}. 

While recent accelerators provide native INT8/INT4/FP8 tensor units and many quantization schemes are explicitly co-designed for such hardware, this makes the inference stack contingent on access to specific GPU generations and high-end datacenter GPUs, which runs counter to our goal of democratizing LLM inference. In contrast, our goal is to design a modest quantization regime that remains effective across commodity single-GPU environments without assuming specialized low-precision compute support, while still reducing expert transfer cost (including metadata overhead) and preserving model quality on standard language modeling benchmarks. This design target motivates the quantization strategy we introduce in Mira (details in Section~\ref{sec:Mira-quant}).


\subsection{Sparse Expert Activation}
\label{sec:sparse-activation}

MoE models rely on sparse expert activation. For each token, a small subset of experts (e.g., top-$k$) is selected by the router, while the remaining experts remain idle. This sparsity is highly skewed and input-dependent as a few experts may be frequently activated for common workloads, while a long tail of experts fires rarely but is essential for maintaining model quality on specific topics or user queries. Figure~\ref{fig:expert_activation} illustrates the expert activation behavior of Mixtral-8x7B \cite{jiang2024mixtralexperts} on four representative workloads: dark-humor detection, movie recommendation, vitamin fact verification, and physics questions from the MMLU dataset \cite{hendrycks2020measuring}. Each panel visualizes a single task: the x-axis enumerates the 32 MoE layers in Mixtral, the y-axis indexes the eight experts within each layer, and the color encodes the fraction of routed tokens in that task assigned to a given expert at that layer (i.e., per-layer activation frequency). Bright cells correspond to hot experts that receive a large share of tokens in that layer, while darker cells correspond to rarely selected experts.

These heatmaps highlight several salient patterns. First, within a task, expert selection is skewed across layers. Most layers are dominated by a few experts, which appear as bright horizontal bands, while the remaining experts in that layer are only weakly activated for certain tokens. Thus, a small subset of experts per layer handles the bulk of traffic, and the others are only sporadically used at a given time during decode phase in local inference. Second, expert activation patterns vary substantially across tasks. For the vitamin-facts dataset, the activation is comparatively more uniform than for the other workloads. Moreover, the set of the hot experts per layer change with the task. These observations underscore that expert popularity is both layer and workload dependent, which in turn motivates adaptive cache management that can respond to evolving expert activation patterns rather than relying on fixed popularity statistics.

Consequently, maintaining a static GPU-resident cache of popular experts, for example, pinning the top experts per layer based on offline profiling, as in static caching schemes such as Fiddler \cite{kamahori2024fiddler}, is insufficient. Static caches assume a stable routing distribution; they cannot adapt when the query mix shifts over time, when the model processes long-context inputs where topics drift within a single sequence, or when a user's local workload deviates from the distribution used to build the cache. As a result, they either waste VRAM on experts that are no longer hot or incur frequent cache misses for newly important experts, leading to PCIe stalls and degraded throughput.

A natural alternative is to react to routing decisions by prefetching experts for the next layer, e.g., an $i \rightarrow i{+}1$ prefetching policy that stages experts predicted to be used in the following MoE layer \cite{hwang2024pre}. While such next-layer prefetching can hide some latency in large-batch, high-throughput settings, it is often inadequate for the single-batch, interactive inference regime that dominates local LLM usage. In the single-batch case, there is limited opportunity to overlap PCIe transfers with useful computation, so any misprediction or late prefetch directly translates into additional end-to-end latency.

When experts are stored in higher-precision formats (e.g., INT8 instead of ultra-low-bit formats), each migration is more expensive, making $i{+}1$ prefetching even less effective, as the system must wait for the top-$k$ experts to arrive to finish computation. Further, inaccurate fetching wastes both bandwidth and scarce VRAM. These limitations motivate a design in which a HOT cache and a STAGE prefetch window work together, with both regions explicitly aware of sparse and dynamic activation patterns.
The HOT region holds a dynamically maintained set of experts that are consistently popular over recent routing history, providing low-latency access for the majority of tokens. The STAGE region, in contrast, is driven by a predictive prefetcher that targets experts that are not currently hot but are likely to be needed soon.

\begin{figure}[t]
    \centering
    \includegraphics[width=0.45\textwidth, scale=1.2]{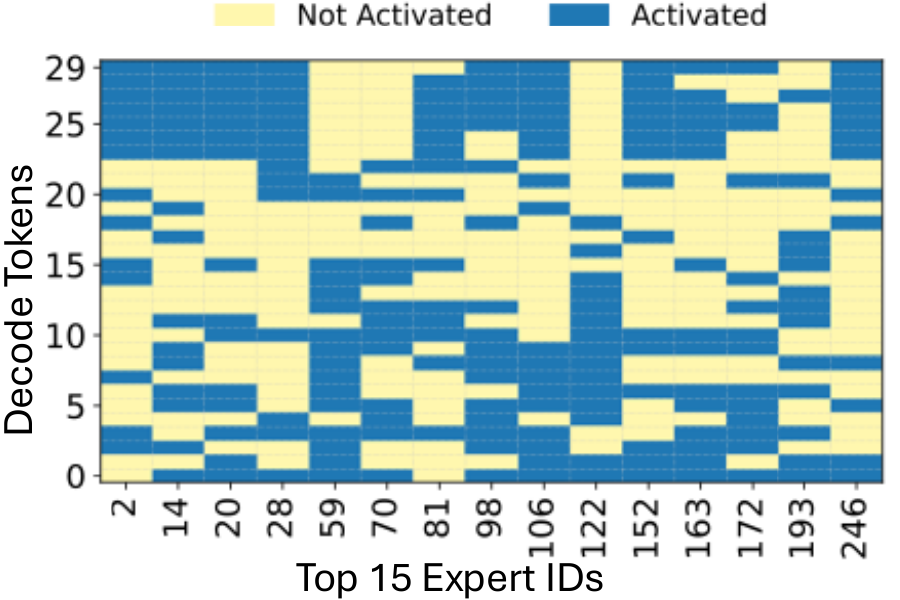}
    \vspace{-12pt}
    \caption{Top-15 expert activations across decode tokens on the movie recommendation dataset using Mixtral-8x7B. Each column tracks the activation frequency of one expert over a decode sequence.}
    \label{fig:decode_tokens}
    \vspace{-10pt}
\end{figure}

Crucially, the HOT set must be continuously rebalanced. In a model like Mixtral-8x7B, which has 8 experts per layer across 32 layers (i.e., 256 experts in total), keeping all experts resident would exceed the capacity of a single commodity GPU, so only a subset of experts can live in GPU memory at any given time, and which experts deserve those slots must change as the workload evolves. Figure~\ref{fig:decode_tokens} provides a zoomed-in view of Figure~\ref{fig:expert_activation}; we focus on the 15 most frequently activated experts on the movie recommendation dataset and track how the hot expert set evolves over decode tokens. Across decode steps, no single expert remains dominant. For example, expert 59 is heavily activated for early tokens but becomes inactive later on; beyond that point, keeping it in the HOT cache provides little benefit. If it is needed again sporadically, it can be brought in via the prefetcher. In contrast, expert 28 is not frequently activated at the beginning but becomes active for a long contiguous span of later tokens, making it a strong candidate to be promoted into the HOT cache during that phase. This figure is a small snapshot of what happens during decoding. Over longer decode phases, expert activations exhibit even richer temporal variation. By combining predictive staging for out-of-the-blue experts with lightweight, online rebalancing of the HOT set, the system can track the evolving sparse activation structure of the MoE, reducing both cold misses and unnecessary transfers while respecting tight VRAM budgets in local inference environments.

\begin{figure*}[t]
\vspace{-0pt}
  \includegraphics[width=2.1\columnwidth, scale=0.5]{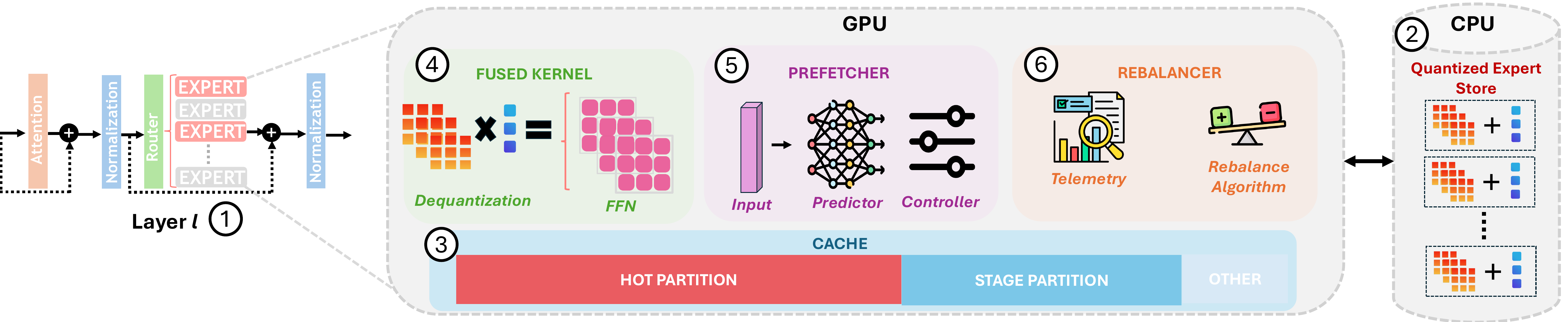}
  \vspace{-20pt}
  \caption{\textbf{Mira's Runtime Components:} At runtime, Mira maintains a two-level GPU expert cache with HOT and STAGE regions. During inference, experts are fetched from this cache and processed using a custom kernel that fuses dequantization with FFN computation. A lightweight predictor forecasts which experts will be needed shortly and triggers asynchronous prefetching into the STAGE region. Meanwhile, token-level telemetry informs a rebalancer that periodically adjusts the HOT expert set for optimal reuse. Offline, on CPU side, all expert FFN weights are converted into a unified INT8 format.}
  \label{fig:high_level}
  \vspace{-10pt}
\end{figure*}

\section{Mira: An Algorithm-System Co-Design}
\label{sec:Mira}

This section presents \textbf{Mira}, our algorithm-system co-designed runtime for MoE transformers. Mira jointly optimizes three components: a custom INT8 quantization layout tailored to expert feed-forward blocks, a two-level GPU cache for expert weights, and a predictive prefetcher with adaptive rebalancing driven by MoE routing signals. Mira is designed to be compatible with any transformer MoE that exposes experts and routing decisions via standard interfaces.

\subsection{Design Objectives and Overview}
\label{sec:Mira-overview}

Mira is designed around three primary goals: \textbf{(1) Model Generality:} The runtime must support a variety of MoE models, avoiding hardcoded assumptions about expert layout, routing scheme, or training procedure. Its interface requires only the ability to enumerate experts, access their weights, and observe router assignments. \textbf{(2) Single-GPU Efficiency:} The system must be VRAM-aware and deliver high performance on a single accelerator, including commodity devices with modest memory capacity, without relying on multi-GPU expert sharding or specialized tensor-core formats. \textbf{(3) Adaptive Self-Management:} Mira must operate as a self-sufficient system that dynamically drives its own caching, prefetching, and memory allocation policies. Our adaptive cache and predictive expert staging are the core mechanisms designed to achieve these goals.

Figure \ref{fig:high_level} conceptually illustrates Mira's runtime components. During sequential layer-by-layer inference, Mira's runtime mainly impacts how the expert FFN is computed \whitecircle{1}. All expert weights are first converted offline into a unified, custom INT8 representation \whitecircle{2}. At runtime, Mira maintains the GPU memory as a two-level expert cache \whitecircle{3}. The HOT region holds resident experts that demonstrate frequent reuse, which can be invoked with no host-to-device traffic. The STAGE region acts as a prefetch window, holding experts expected to be needed in the near future based on predictor output.

During computation, experts are fetched from this cache. Mira employs a custom kernel that fuses the dequantization of INT8 weights with the FFN computation itself \whitecircle{4}. To facilitate this prefetching, Mira attaches a small predictor head to each applicable MoE layer ($i$). This head consumes hidden states and router statistics from layer $i$ to predict which experts will be required two layers ahead ($i+2$) \whitecircle{5}. These predictions initiate asynchronous data transfers from the CPU-pinned store into the STAGE region. Concurrently, Mira continuously gathers token-level telemetry, quantifying expert assignments, data source tiers (HOT, STAGE, or CPU), and cache miss rates. This data feeds a rebalancer that adaptively optimizes the HOT set composition over time \whitecircle{6}.

The design is intentionally non-invasive relative to the base transformer model. Components like embedding layers, self-attention sublayers, layer norms, and the final LM head remain in BF16 on the GPU and execute unchanged. Only the expert FFNs are wrapped by Mira's runtime: when the model reaches an MoE layer and calls its experts, that call is redirected into Mira's execution pipeline, which handles quantized storage, cache lookup, and dequantization. As a result, the interface seen by the rest of the model, particularly the shapes and semantics of hidden states and output logits, remains identical to the baseline.

\subsection{Quantization-Aware Expert Store}
\label{sec:Mira-quant}

Mira's custom quantization representation targets MoE architecture where each expert is a three-matrix feed-forward block with an up projection \(W_1\), a down projection \(W_2\), and a gate modulation matrix \(W_3\) with the up and gate matrices \(W_1\) and \(W_3\) have shape \(R \times C\), while the down matrix \(W_2\) has shape \(C \times R\), where \(R\) is the expert hidden dimension and \(C\) is the model hidden dimension.

Mira converts each expert into a compact INT8 format. The up and gate matrices are stored as INT8 tensors of shape \(R \times C\), and the down matrix is stored as an INT8 tensor of shape \(C \times R\). A single BF16 vector \(s \in \mathbb{R}^R\) stores per-row scales that are shared across all three matrices. The quantization is constructed as follows. For each row index \(r\), Mira computes
\[
\alpha_r = \max\left(
\lVert W_1[r,:] \rVert_\infty,\;
\lVert W_3[r,:] \rVert_\infty,\;
\lVert W_2[:,r] \rVert_\infty
\right),
\]
and defines a scale
\[
s_r = \frac{\max(\alpha_r, \epsilon)}{127},
\]
where \(\epsilon\) is a small constant to avoid division by zero. The three matrices are then quantized using this shared per-row scale:
\begin{align}
\hat{W}_1[r,c] &= \mathrm{round}\!\left(\frac{W_1[r,c]}{s_r}\right),\\
\hat{W}_3[r,c] &= \mathrm{round}\!\left(\frac{W_3[r,c]}{s_r}\right),\\
\hat{W}_2[c,r] &= \mathrm{round}\!\left(\frac{W_2[c,r]}{s_r}\right),
\end{align}
and each value is clamped to the range \([-127,127]\) so it can be stored in a signed 8-bit integer. Our custom INT8 scheme employs joint compression, utilizing a single scale factor shared across all three matrices, unlike standard per-matrix INT8 quantization. As shown in Table \ref{tab:quantization_perplexity} and Section \ref{subsec:accuracy}, this approach incurs a negligible impact on model quality compared to the standard INT8 baseline. 

The primary benefit of our shared-scale INT8 format is reducing per-expert staging overhead in MoE inference, where experts are repeatedly migrated and prefetched under limited GPU memory. Standard per-row INT8 stores one scale vector per matrix, requiring multiple scale transfers and additional copy operations each time an expert is staged. In contrast, our scheme uses a single shared scale per expert, reducing scale metadata movement and the number of CPU $\rightarrow$ GPU copy operations during staging. Although end-to-end staging time is typically dominated by transferring the INT8 weight matrices, these metadata and copy-call savings accumulate when staging multiple experts (e.g., during prefetch) and reduce transfer-side overhead. In a scale-only microbenchmark, shared scaling reduces CPU $\rightarrow$ GPU scale transfer latency by 1.64$\times$ on Mixtral-8x7B, by 1.62$\times$ on Mixtral-8x22B, and by 1.68$\times$ on DeepSeek-V2-Lite.


This INT8 representation is deliberately moderate. We avoid ultra-low-bit schemes such as INT3 or INT2, which require elaborate packing, additional dequantization metadata, and often incur noticeable accuracy loss. INT8 offers a pragmatic compromise: it reduces the per-expert footprint sufficiently to make large-scale caching and offloading attractive, yet it remains conceptually simple, hardware-portable, and amenable to efficient dequantization strategies.

From an implementation perspective, our joint quantization scheme's only firm prerequisite is that expert FFNs are expressible as a three-matrix (up, down, gate) transformation. This pattern is characteristic of a broad class of contemporary MoE architectures, making our technique widely applicable. Furthermore, Mira's runtime is modular: if an MoE model deviates from this structure, it can employ a standard quantization technique while still leveraging Mira's core inference components. 

\subsection{HOT+STAGE Expert Cache in GPU Memory}
\label{sec:Mira-cache}

\begin{figure}[t]
    \centering
    \includegraphics[width=0.5\textwidth, scale=1.2]{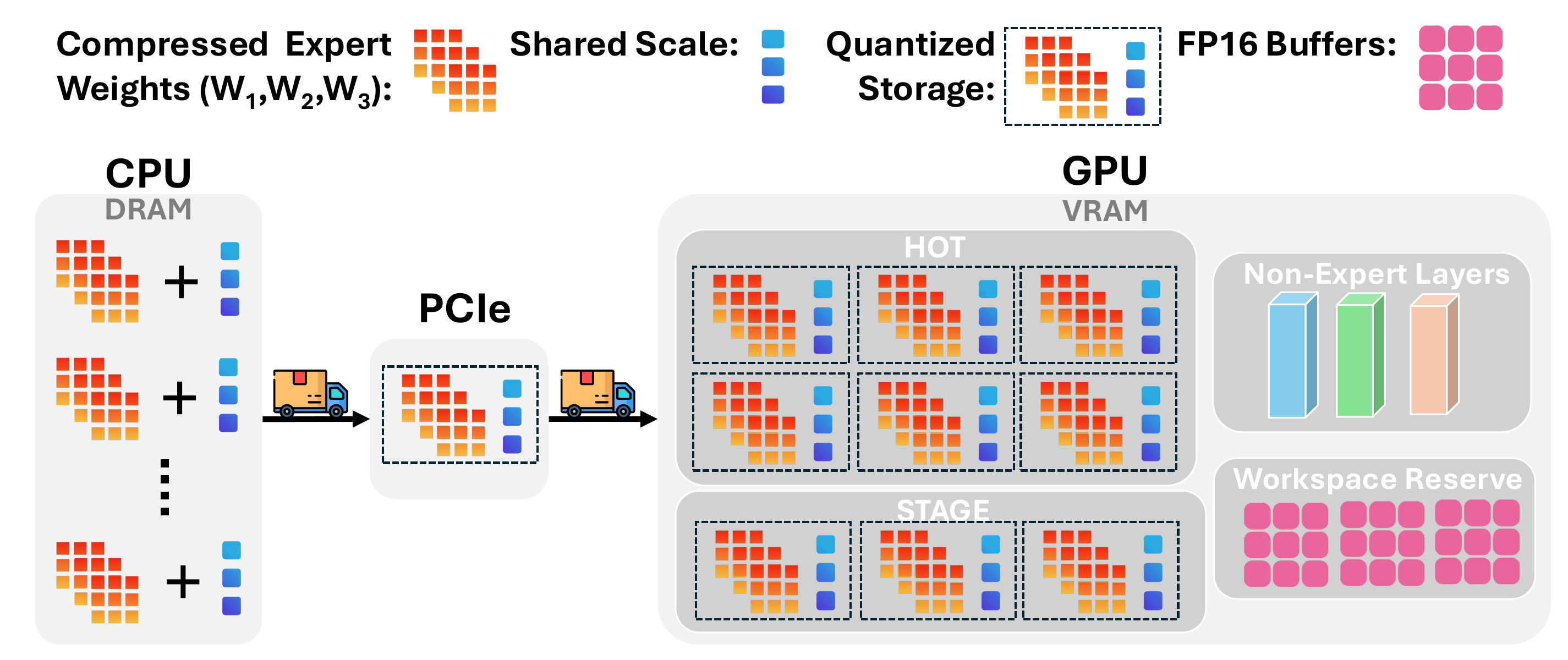}
    \vspace{-12pt}
    \caption{Mira’s memory layout. Compressed expert weights $W_1$, $W_2$, and $W_3$ are combined (denoted by $+$) on the CPU. The weights and a shared scale are packed (indicated by the dashed line) and transferred to the GPU.}
    \label{fig:memory_hiearchy}
    \vspace{-10pt}
\end{figure}

At runtime, Mira manages GPU memory dedicated to experts according to an explicit byte-level budget. Mira estimates how much capacity must be reserved for non-expert model layers and BF16 working buffers (workspace reserve) used during. The remainder is available for caching expert weights. This remaining budget is expressed in units of expert slots. Since the compressed expert size is known, Mira can compute how many experts can be resident simultaneously. It then partitions the budget into two regions dedicated to HOT and STAGE, respectively. The partitioning is dynamic and depends on whether the runtime is operating under tight memory constraints. Figure \ref{fig:memory_hiearchy} illustrates this memory layout.

The HOT partition in Mira contains experts that are fully resident as compressed INT8 tensors in GPU memory and are expected to exhibit long-term reuse. They remain in the cache until evicted by the rebalancing policy. The STAGE partition acts as a prefetch window. It holds experts that have been copied to GPU memory in anticipation of near-future use, because the predictor has identified them as likely to be activated in upcoming layers. STAGE entries are short-lived; they are dropped when their target layer falls sufficiently behind the layer currently being processed. 


Mira adapts its cache behavior to the available VRAM budget, operating in either a small-VRAM or large-VRAM mode. This distinction primarily governs the provisioning and eviction policies for the STAGE partition. In large-VRAM mode, the STAGE partition is provisioned generously. Experts are granted longer residency, and eviction policies are less aggressive. This larger capacity provides a longer observation window, allowing the runtime to more accurately identify frequently demanded experts for eventual promotion to the HOT cache. Conversely, the small-VRAM mode operates under an extremely tight budget where only a small percentage of experts can be GPU-resident, making the STAGE partition's capacity minimal and not intended for long-term residency.

In this small-VRAM mode, the STAGE partition is provisioned to hold only a handful of predicted experts for immediate future layers (e.g., layer $i+2$ while processing layer $i$), and its memory is aggressively reclaimed post-computation. If the STAGE partition is full when a new expert is predicted, the runtime declines to stage that expert, defaulting to the CPU path when that expert is eventually required. This design strictly conserves VRAM while still allowing low-capacity GPUs to exploit predictive staging and overlap whenever the STAGE partition can accommodate it. 

This aggressive eviction makes efficient HOT set management paramount, and we found that Mira's rebalancer provides significant benefits over standard cache policies such as Least Recently Used (LRU) or Least Frequently Used (LFU). A conventional LRU policy is suboptimal, as it promotes staged experts based on recency of use, not frequency, which pollutes the HOT cache with potential cold experts that may not be subsequently reused. A pure LFU policy is also not optimal, as it suffers from cache pollution (where old, popular experts never get evicted). Mira's rebalancer, in contrast, acts as a history-based gatekeeper. It uses the STAGE partition as a probationary area, maintaining comprehensive usage statistics for all experts in both STAGE and HOT partitions. An expert is only promoted from STAGE after the rebalancer observes a persistent history of frequent access. This check ensures that only experts truly worthy of the high-value HOT cache capacity are promoted, avoiding issues with standard caching policies.

\subsection{Adaptive HOT Cache Rebalancing}
\label{sec:Mira-rebalance}

To effectively determine HOT residency, Mira maintains fine-grained, token-level accounting. For each expert, the runtime tracks two key statistics: (1) the total number of tokens routed to it by the MoE router, and (2) the source memory tier for those tokens (i.e., served from HOT, STAGE, or CPU). These counters are maintained over a sliding window of tokens and are reset after each rebalancing interval.

From these raw counts, the runtime computes higher-level metrics by categorizing the service path for each token. A HOT-served token is one processed by an expert already resident in the high-performance HOT cache, which we use to measure cache utilization. A STAGE-served token is one processed by an expert that was successfully pre-fetched into the STAGE partition, which measures prediction effectiveness. Finally, a CPU-served token represents a complete cache miss where the expert had to be fetched from CPU memory upon demand, which directly quantifies the cost-of-miss.

This accounting provides a concrete link between abstract MoE routing decisions and their physical system costs. For example, an expert with a high routed volume but also a high CPU-served count incurs significant PCIe and dequantization overhead, making it a strong promotion candidate. Conversely, a HOT-resident expert that serves few tokens (i.e., low HOT-served count relative to its peers) represents poor VRAM utilization and becomes a candidate for eviction.

After processing a predetermined number of tokens, Mira triggers its rebalancing pass, which operates in three phases. 

\begin{itemize}
    \item  \textbf{Victim Scoring}: The rebalancer examines each HOT-resident expert and assigns a victim score. Experts with low routed volume in the current window receive high scores, marking them for eviction. This score also incorporate cache-level metadata, such as recency and frequency
    \item \textbf{Candidate Scoring}: Concurrently, it considers all non-HOT experts with recent traffic and assigns a candidate score. This score prioritizes experts with high routed volume and a high number of CPU-served tokens, indicating both high demand and a high cost-of-miss
    \item \textbf{Bounded Swapping}: Finally, the rebalancer performs a bounded number of swaps, evicting the worst-scoring HOT residents and promoting the best-scoring candidates. The number of swaps per interval is intentionally limited to prevent cache thrashing. The token threshold to trigger balancing and the number of swaps are offline-tuned hyperparameters, derived from a performance sweep to find the optimal trade-off between rebalancing overhead and cache adaptability for a given MoE architecture.
    
\end{itemize}

This rebalancing procedure is fundamentally MoE-aware in a way that generic caching policies (e.g., LRU/LFU) are not. It does not treat experts as anonymous, uniformly-sized cache lines. Instead, it leverages the semantic signal of MoE routing decisions and the provenance of served tokens (i.e., HOT, STAGE, or CPU) to make cost-aware decisions about which experts have earned residency.

\subsection{Predictive Expert Staging}
\label{sec:Mira-predict}

Adaptive caching alone is insufficient to hide expert offload latency on a single GPU. To move beyond this limitation, Mira uses per-layer predictor heads that attempt to forecast expert usage a few layers ahead. For each MoE layer \(i\), except the last two, we attach a small neural module that produces a probability distribution over experts in layer \(i+2\).

The predictor consumes a compact feature vector constructed from the state at layer \(i\). Specifically, Mira mean-pools the hidden activations across tokens to obtain a single vector summarizing the current semantic context, averages the router logits to capture the marginal tendency of the router to favor each expert, and builds a histogram over the experts actually selected at that layer. In addition, Mira identifies the top-$k$ experts who receive the most tokens at layer \(i\). These features together encode both the content being processed and the router's current behavior in an inexpensive-to-compute form.

The predictor head itself, which we call a MixtureHead, is implemented as a dynamic mixture of two sub-models. The first branch, an ActHead, processes the semantic context and router logits. The second branch, a HistHead, processes the historical expert selection features. These two branches produce separate logit predictions, which are then combined using a contextual gate. This small gating network learns to output a dynamic weight, $\alpha$, that adaptively balances the influence of the activation-based and history-based branches for each prediction. 


Training of the predictor heads is entirely offline and does not modify the base model. The runtime collects traces in which it records the feature vector at layer \(i\) and the actual expert activations at layer \(i+2\) for a representative workload. Each predictor is trained with a standard cross-entropy loss to align its output distribution with the empirical distribution of routed tokens across experts in the target layer. During inference, the trained predictors are loaded to the GPU and run in evaluation mode, adding minimal computational overhead relative to the main expert FFNs.

During inference, for each expert predicted for layer \(i+2\), Mira checks if the expert is already present in HOT or STAGE partition, the runtime records a prefetch skipped event and moves on. If the expert is not yet cached, the prefetch controller attempts to admit a new STAGE entry. Admission triggers an asynchronous copy of the compressed expert from the CPU-pinned store to GPU memory on a high-priority CUDA stream. The controller also records that expert \(e\) is staged for target layer \(i+2\) so that repeated staging requests for the same layer do not issue redundant copies.

\begin{figure}[t]
    \centering
    \includegraphics[width=0.5\textwidth, scale=1.2]{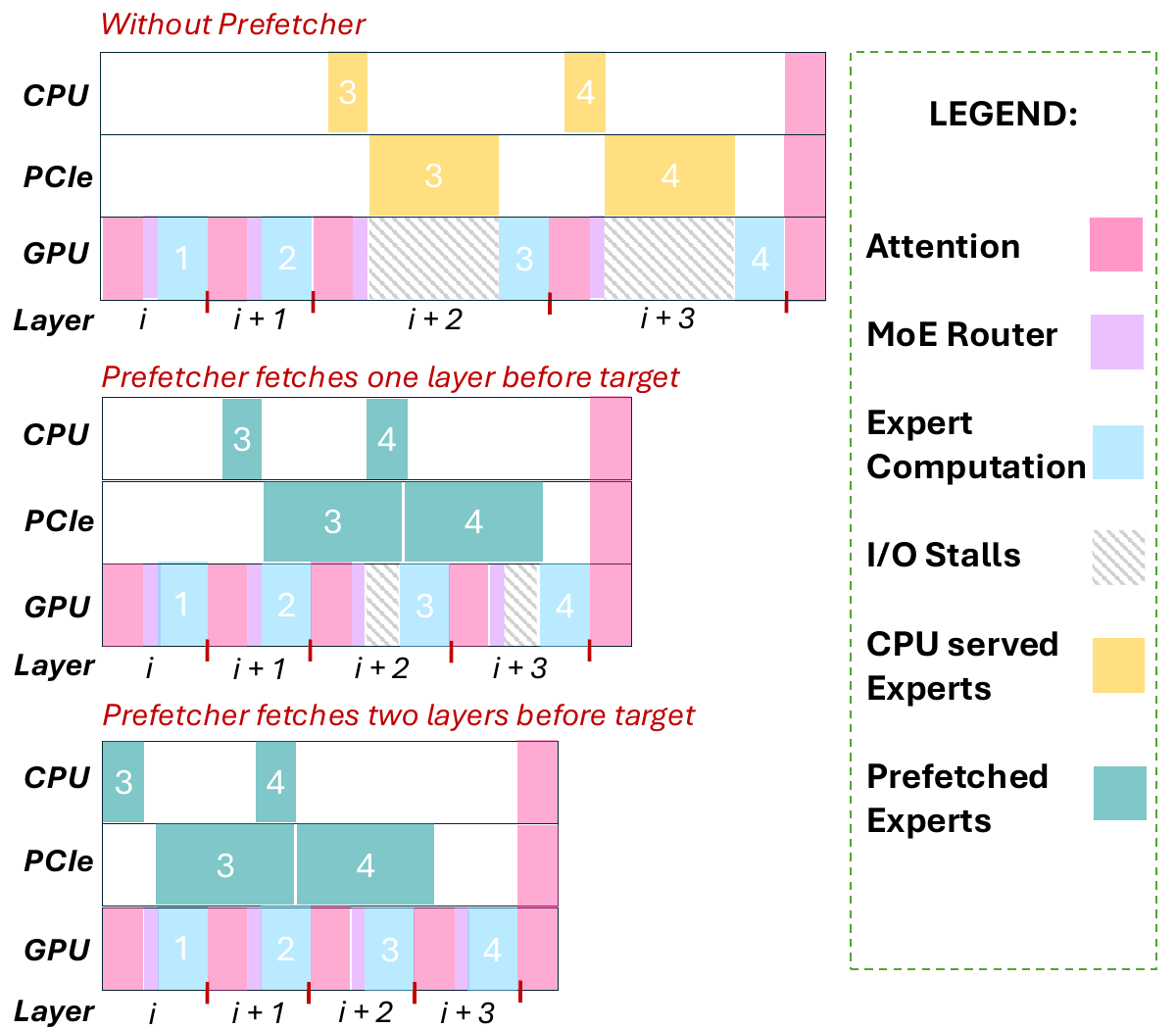}
    \vspace{-12pt}
    \caption{Timeline illustration of expert computation under varying hardware setups.}
    \label{fig:prefetch_pipeline}
    \vspace{-10pt}
\end{figure}

Our choice of a two-layer lookahead (predicting for layer \(i+2\) at layer \(i\)) is dictated by the system's latency profile as illustrated in figure \ref{fig:prefetch_pipeline}. Even with INT8 compression, the end-to-end latency of fetching an expert, including the PCIe transfer, on-device dequantization to BF16, and the predictor's own inference overhead, is non-trivial. The \(i+2\) target provides a sufficiently large window to overlap this data transfer with the active computation of layers \(i\) and \(i+1\). This \(i+2\) window represents a critical sweet spot. We experimentally implemented and evaluated a one-layer lookahead ($i+1$) predictor, but found it was insufficient to fully hide this latency, especially in single-token inference. Conversely, looking further ahead would increase prediction uncertainty and risk wasting bandwidth. Empirically, our \(i+2\) predictor provided the necessary slack time while, importantly, achieving the same prediction accuracy as the \(i+1\) model, indicating that sufficient predictive signals exist two layers deep.

\subsection{Algorithm-System Co-Design}
\label{sec:Mira-codesign}

Mira derives its effectiveness from the tight coupling between algorithmic insight and systems design. On the algorithmic side, it introduces per-layer predictors trained on real routing traces and a custom quantization scheme tuned to the expert FFN structure. These components exploit MoE-specific patterns while leaving the base model and router unmodified. On the systems side, Mira provides a two-level (HOT+STAGE) GPU cache with explicit VRAM budgeting, a predictive prefetcher that acts on the predictor's signals, and utilization-aware policy for HOT rebalancing.

This co-design is exemplified by the expert execution pipeline, which is carefully structured to manage the overhead of our custom quantization. To keep this overhead manageable, Mira pre-allocates BF16 working buffers on the GPU for the dequantized weights and scales. These buffers are reused across experts and decoding steps to eliminate allocator churn on the critical path. The runtime then orchestrates execution using two CUDA streams: a dequantization stream copies INT8 data and BF16 scales (initiating H2D transfers from pinned memory if needed), and a compute stream performs the FFN computation. 
The expert's FFN computation is fused with weight dequantization, avoiding the need to materialize the full-precision weights in VRAM. 

Neither the algorithmic nor the systems components would be as effective in isolation. Predictors are inert without a cache system to act on their signals, and a naive cache, blind to MoE routing structure, would struggle to anticipate expert demands under tight VRAM budgets. By designing both pieces together, Mira turns the statistical regularities of expert usage into concrete reductions in PCIe stalls and improved effective throughput, all while maintaining model quality. In Section \ref{sec:eval}, we quantify these benefits and show that Mira enables larger MoE models to be served within constrained VRAM budgets.

\begin{figure*}[t]
\vspace{-0pt}
  \includegraphics[width=2.1\columnwidth, scale=1.2]{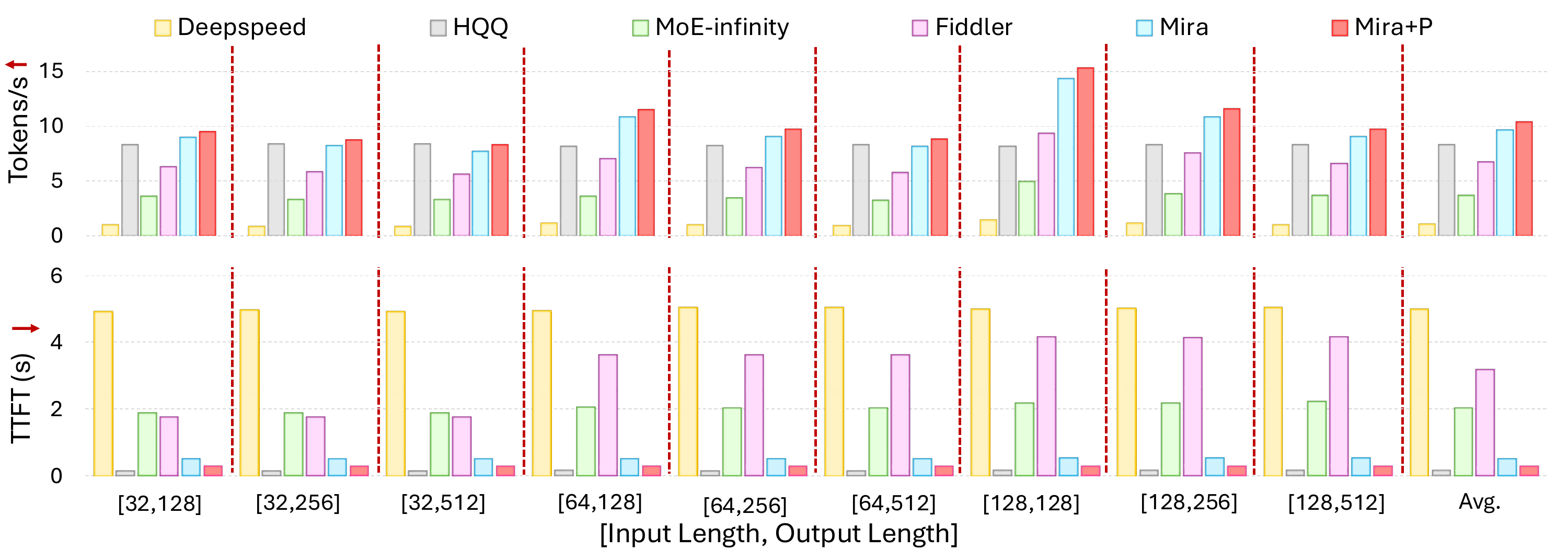}
  \vspace{-20pt}
  \caption{Inference performance comparison of Mira+P (with prefetcher) and Mira (without prefetcher against baselines in Environment 1 (48GB VRAM), measured by Time-to-First-Token (TTFT) and throughput (Tokens/s) for varying input and output token lengths. \emph{ Higher is better for Tokens/s; lower is better for TTFT.}}
  \label{fig:env1_perf}
  \vspace{-10pt}
\end{figure*}

\section{Experimental Setup}
\label{sec:experimental_setup}

Our experiments are designed to achieve several key objectives. First, we benchmark Mira's inference performance against state-of-the-art baselines that specialize in expert offloading, compression, or hybrid approaches. Second, we evaluate Mira's task accuracy. Third, we demonstrate Mira's robustness by assessing its performance across different hardware setups with varying VRAM capacities. 

\textbf{Baselines:} We employ following state-of-the-art baselines for single-GPU inference on memory-constrained systems: DeepSpeed-MII \cite{deepspeed}, an inference system that utilizes ZeRO-Infinity \cite{rajbhandari2021zero} to offload expert parameters to CPU and NVMe; Mixtral-Offloading \cite{eliseev2023fast}, an MoE framework that integrates expert prediction and caching mechanisms with 4-bit quantization; MoE-Infinity \cite{xue2025moeinfinityefficientmoeinference}, an offloading system that uses sequence-level activation tracing to perform sparsity-aware expert prefetching and caching; Fiddler \cite{kamahori2024fiddler}, a CPU-GPU co-execution system that minimizes data transfer overhead by dynamically offloading portions of the expert computation to the CPU;  and HQQ, a 2-bit weight-only quantization with 128 group size  \cite{badri2023hqq}.

\textbf{Datasets:} To evaluate inference performance, we use ShareGPT \cite{zheng2023judging}, a dataset of human-chatbot conversations chosen to model realistic expert selection patterns, randomly sampling conversations with varying input and output lengths; this setup is similar to the Fiddler~\cite{kamahori2024fiddler} and FloE~\cite{zhou2025floe}. 

Our comparative analysis focuses mainly on Mixtral-8x7B \cite{jiang2024mixtralexperts}, as it is the only model supported across the open-sourced codebases of key baselines Fiddler \cite{kamahori2024fiddler} and Mixtral Offloading \cite{eliseev2023fast}, but also includes DeepSeek-V2-Lite-Chat to demonstrate Mira's generalizability. 

\textbf{Metrics:} For a comprehensive performance view, we report two key metrics: Time-To-First-Token (TTFT) to measure initial latency, which is especially important for interactive uses cases and end to end tokens-per-second to measure throughput of the system. We also report accuracy on downstream tasks performance using lm-eval-harness \cite{eval-harness} by sampling from various datasets.

\textbf{System configurations:} We evaluated Mira on three distinct hardware environments to test performance across different VRAM capacities and memory bandwidths. \textbf{Environment 1} featured an NVIDIA RTX 6000 Ada (48GB VRAM, 960 GB/s bandwidth) paired with an Intel(R) Xeon(R) w5-2565X. \textbf{Environment 2} used an NVIDIA RTX A5000 (24GB VRAM, 768 GB/s bandwidth) with an AMD EPYC 7352. \textbf{Environment 3} consists of NVIDIA RTX 3070 (8GB VRAM, 448 GB/s bandwidth) with an Intel(R) Xeon(R) w5-2565X. The number of processor cores is restricted to 8 to mirror resource-constrained environments.

\section{Evaluation}
\label{sec:eval}

\begin{figure*}[t]
    \centering
    \includegraphics[width=2.1\columnwidth, scale=0.5]{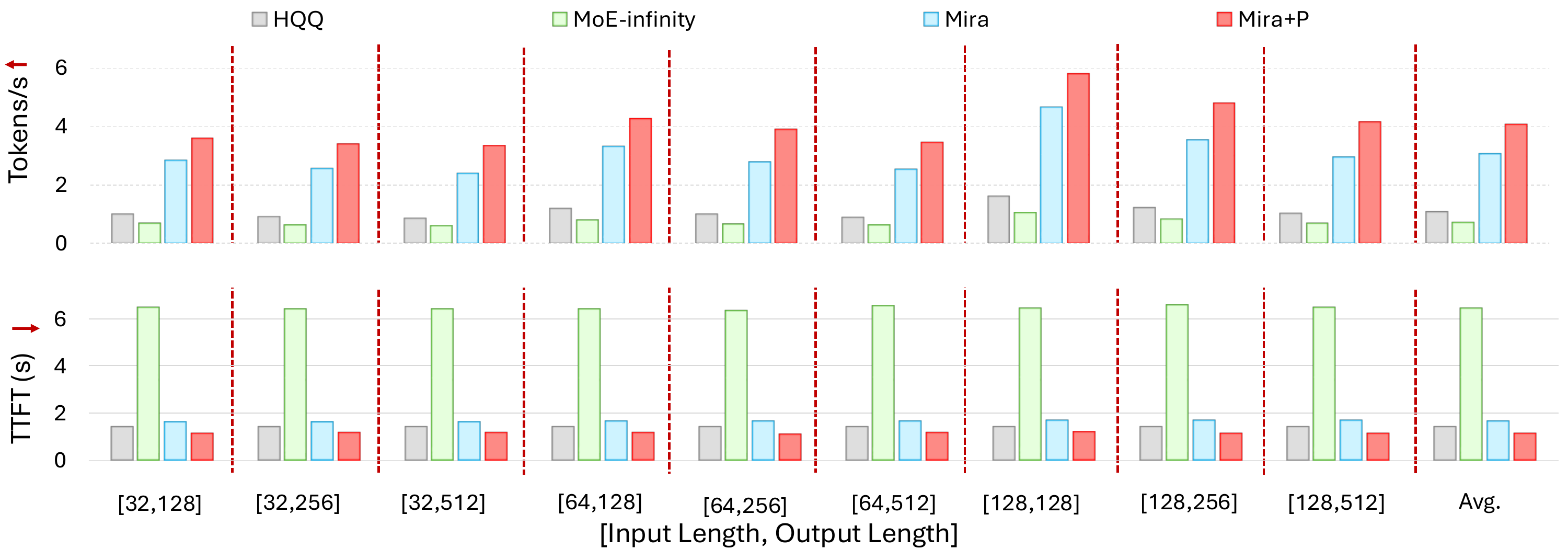}
    \vspace{-12pt}
    \caption{Inference performance comparison of Mira+P (with prefetcher) and Mira (without prefetcher against baselines in Environment 2 (24GB VRAM), measured by Time-to-First-Token (TTFT) and throughput (Tokens/s) for varying input and output token lengths. \emph{ Higher is better for Tokens/s; lower is better for TTFT.}}
    \label{fig:env_perf2}
    \vspace{-10pt}
\end{figure*}

\subsection{Performance}
\label{sec:perf}

Figure \ref{fig:env1_perf} illustrates the performance of Mira compared to the baselines in Environment 1, which features 48GB of VRAM. In terms of end-to-end throughput (TPS) averaged across different input and output lengths, Mira+P (with prefetching enabled) achieves an average speedup of 10.46$\times$ over DeepSpeed, 1.25$\times$ over HQQ, 2.86$\times$ over MoE-Infinity and 1.55$\times$ over Fiddler. These end-to-end time reports for Mira include the overhead of our predictor and rebalancer, and we observe that these additional runtime components do not have a significant performance impact. The speedup over Mira (prefetching disabled) is 1.07$\times$. This modest gain is expected, as the 48GB VRAM allows the HOT cache to hold a large number of experts. Consequently, the prefetcher has a limited role during decoding because most predicted experts are already resident in the cache, and the few misses do not contend for I/O. This memory advantage also benefits other baselines; Fiddler sees better performance as fewer computations are offloaded to the CPU, and The HQQ baseline also fits fully on-device, but its performance is bound by the computational overhead of dequantizing its highly aggressive quantization scheme.

The performance advantages are more pronounced in the Time-to-First-Token (TTFT) metric, which is critical for interactive sessions. Here, Mira+P achieves an average speedup of 18.38$\times$ over DeepSpeed, 11.71$\times$ over Fiddler, 7.49$\times$ over MoE-Infinity, and 3.62$\times$ over Mixtral Offloading. The HQQ baseline also has low prefill overhead, because the entire model is fully resident in GPU VRAM, eliminating any I/O latency. Notably, Mira+P demonstrates a 1.85$\times$ speedup over Mira for TTFT. This significant gain is because the prefill stage processes multiple tokens per layer, activating a far greater number of experts than in the single-token decode stage. The prefetcher excels in this scenario, correctly anticipating the diverse expert demand and ensuring those weights are staged on the GPU in time for computation, thus effectively hiding I/O latency.

Figure \ref{fig:env_perf2} presents the performance results for Environment 2, which is significantly more memory-constrained with 24GB of VRAM (half of Environment 1). In this environment, we had to omit two baselines. DeepSpeed was excluded as it required an excessive amount of CPU RAM for offloading, which was unavailable in this environment and it was slow in general. This observation aligns with prior work \cite{kamahori2024fiddler, zhou2025floe}, which also noted DeepSpeed sub-optimal for such memory-constrained inference. We also omitted Fiddler, as its performance is critically dependent on the AVX512 instruction set, which the CPU in this environment lacks. This hardware mismatch resulted in extremely poor performance (we observed a 21.81$\times$ average speedup over it), making a direct comparison misleading. This highlights a key advantage of Mira as it imposes no such limiting hardware requirements, supporting our goal of democratizing LLM access.

In terms of end-to-end throughput, Mira+P achieves a 5.71$\times$ average speedup over MoE-Infinity, 3.86$\times$ over HQQ, and a 1.32$\times$ speedup over Mira.  For the TTFT metric, Mira+P demonstrates advantages, achieving a 5.61$\times$ over MoE-Infinity,  1.23$\times$ over HQQ and 1.42$\times$ over Mira-Base. The speedup over Fiddler was 46.9$\times$. These TTFT gains of Mira+P, particularly over Mira, are directly attributable to the prefetcher, which effectively hides I/O latency that becomes a critical bottleneck in this memory-constrained setting.

\subsection{Accuracy Analysis}
\label{subsec:accuracy}

\begin{table}[h]
    \centering
    \caption{Accuracy comparison on downstream tasks}
    \label{tab:acc_table}
    \resizebox{\columnwidth}{!}{%
    \begin{tabular}{lccccccccc}
        \toprule
        & \textbf{Arc-E} &  \textbf{SCIQ} 
        & \textbf{Winogrande} & \textbf{TriviaQA}\\
        \midrule
        \textbf{FP16}
        & 0.870 & 0.970 & 0.760 & 0.57 \\
        \textbf{Mira}
        & 0.871 & 0.975 & 0.750 & 0.54\\
        \bottomrule
    \end{tabular}%
    }
\end{table}

Table \ref{tab:acc_table} compares the accuracy results of Mira with the FP16 baseline across various downstream tasks. More aggressive quantization systems, such as FloE \cite{zhou2025floe}, have reported notable accuracy degradation on downstream tasks, around 4–7\%. In contrast, INT8 quantization \cite{dettmers2022gpt3} is a well-established and robust technique. When paired with our custom method, it avoids the significant precision loss and outlier-clipping issues that plague lower-bit formats.

We designed our custom INT8 scheme to strike a balance. It provides sufficient model compression to reduce the PCIe data transfer payload and mitigate offloading costs, but does so without incurring the substantial accuracy degradation of more aggressive methods. This sweet spot ensures high model fidelity, which is critical for a practical single-GPU inference system, while still yielding tangible performance benefits from reduced data movement.

\begin{figure}[t]
    \centering
    \includegraphics[width=0.5\textwidth, scale=1.2]{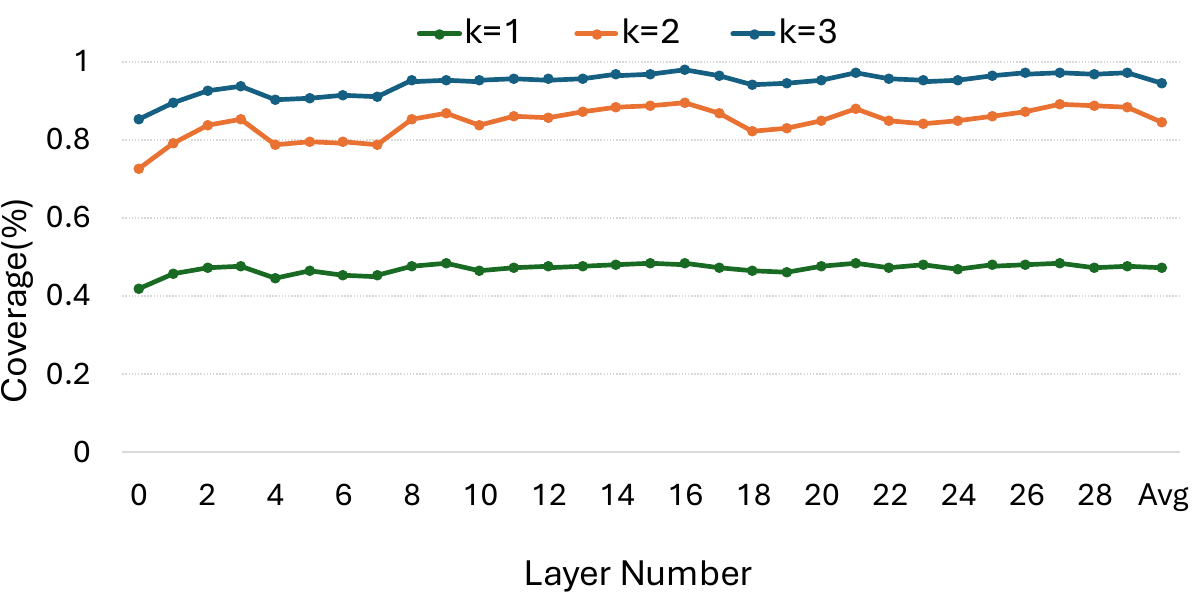}
    \vspace{-12pt}
    \caption{Mass-weighted Coverage@k by source layer for Mixtral-8x7B. The predictor selects the top-k experts to prefetch for a future MoE layer; Coverage@k reports the fraction of routed token mass that lands on those prefetched experts. Higher coverage indicates more effective prefetching.}
    \label{fig:prefetch_acc}
    \vspace{-10pt}
\end{figure}

In Figure \ref{fig:prefetch_acc}, we evaluate how well the predictor supports expert prefetching in Mixtral-8x7B using mass-weighted Coverage@k. For each source layer on the x-axis, the predictor outputs a ranked list of experts for a future MoE layer (at offset $\Delta$); we then measure what fraction of the actual routed token mass at that target layer falls within the predicted top-$k$ experts. Interpreting 
$k$ as the prefetch budget, Coverage@1 corresponds to prefetching a single expert, while Coverage@2 and Coverage@3 correspond to prefetching the top two or three experts, respectively. The consistently higher values for larger 
$k$ show that a small prefetch budget captures most of the routing traffic, indicating that the predictor concentrates probability mass on the experts that dominate routing and would therefore reduce expert stalls and improve cache effectiveness.

\subsection{Additional Results}
\label{subsec:add_results}

\begin{figure}[t]
    \centering
    \includegraphics[width=0.4\textwidth, scale=0.05]{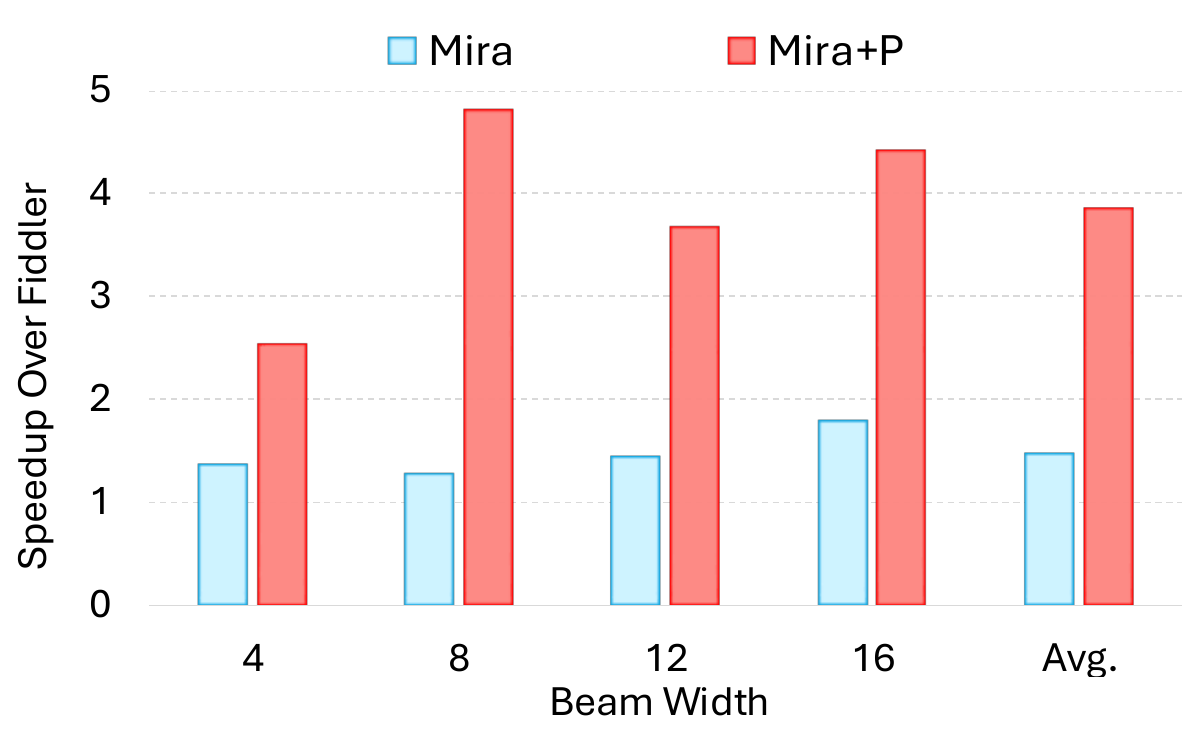}
    \vspace{-12pt}
    \caption{Mira's performance gain over Fiddler with beam width scaling.}
    \label{fig:beam}
    \vspace{-10pt}
\end{figure}

Figure \ref{fig:beam} illustrates how Mira's performance changes as we increase the beam width in beam search decoding. In our earlier experiments, we effectively used greedy decoding (beam width 1). Fiddler also analyzes larger beam widths to evaluate how well the system sustains high-quality generation via beam search under resource-constrained settings. In the single-batch setting, increasing the beam width increases the number of candidate sequences decoded in parallel, stressing the system similarly to a larger effective batch while aiming for better-quality outputs. 

Following Fiddler’s setup, we fix the input and output lengths to 64 and 256 tokens and sweep the beam width. While Fiddler reports its performance gain over llama.cpp in terms of end-to-end tokens per second, here we report our performance gain over Fiddler. We show that both Mira and Mira+P outperform Fiddler as the beam width increases. On average Mira achieves a 1.45$\times$ speedup and Mira+P achieves a 3.84$\times$ speedup. These results demonstrate that Mira scales well in beam-search settings that target higher-quality generation.

\begin{table}[t]
  \centering
  \caption{Mira and Mira+P throughput across varying VRAM budgets.}
  \label{tab:Mira_vram}
  \begin{tabular}{rccc}
    \toprule
    VRAM (GB) & Mira \texttt{(Tokens/s)} & Mira+P \texttt{(Tokens/s)} & Speedup \\
    \midrule
    16 & 1.70 & 2.53 & 1.49 \\
    24 & 2.77 & 3.87 & 1.40 \\
    32 & 3.65 & 4.48 & 1.23 \\
    40 & 5.26 & 7.29 & 1.39 \\
    48 & 8.97 & 9.66 & 1.08 \\
    \bottomrule
  \end{tabular}
\end{table}

Lastly, to isolate and quantify the prefetcher's benefit under varying memory constraints, Table \ref{tab:Mira_vram} presents an ablation study. This experiment was conducted in Environment 1, but we artificially limited the VRAM budget available to the runtime to simulate low-memory scenarios. We used a fixed sequence length (64 input, 256 output). The results support our main findings in Figures \ref{fig:env1_perf} and \ref{fig:env_perf2}, demonstrating that the performance gain from prefetching increases as the VRAM budget decreases. With a smaller budget, the HOT cache shrinks, making cache misses more frequent and costly. The prefetcher's ability to correctly anticipate and stage experts is therefore critical to mitigating this I/O bottleneck.

\section{Related Work}

Recent MoE inference systems for commodity hardware, such as MoE-Infinity \cite{xue2025moeinfinityefficientmoeinference}, Fiddler \cite{kamahori2024fiddler}, FloE \cite{zhou2025floe}, and Mixtral-Offloading \cite{eliseev2023fast}, utilize expert offloading, compression, and basic caching heuristics. Orthogonal works, like Pre-gated MoE \cite{hwang2024pre}, modify the MoE architecture itself to decouple selection from execution. In contrast, Mira retains the original architecture and introduces a predictive, token-level expert cache. It combines a dedicated INT8 representation with a HOT+STAGE cache, driven by a novel $i \to i+2$ expert predictor. Mira thus acts as a drop-in cache manager whose policies are guided by dynamic routing statistics, not static popularity or generic heuristics.

Parallel work on dense LLMs, independent of MoE routing, also addresses memory pressure. Systems like FlexGen \cite{flexgen} and DeepSpeed Zero-Inference \cite{acceleratingllama3fp8} offload dense weights, while vLLM \cite{kwon2023efficient} optimizes the KV. Similarly, post-training quantization methods \cite{lin2024awq,frantar2023qmoe,badri2023hqq,fu2025eaquant,yao2022zeroquant,xiao2023smoothquant,kim2023mixturequantizedexpertsmoqe} uniformly compress all model weights, agnostic to MoE routing. Mira’s design is complementary. It applies a moderate, expert-only INT8 scheme and, crucially, exploits MoE-specific routing signals to drive its dynamic cache placement and prefetching policies.

\section{Conclusions}

This work introduced Mira, an algorithm-system co-design whose effectiveness stems from its proactive, MoE-aware architecture. By introducing a lightweight predictor that forecasts expert usage two layers in advance, Mira transitions from a reactive to a predictive stance, staging experts onto the GPU before they are needed. This predictive prefetching, combined with a novel, telemetry-driven HOT+STAGE cache and an accuracy-preserving INT8 representation, allows Mira to effectively hide I/O latency without compromising model integrity.

By coupling this predictive staging with an intelligent, adaptive runtime, Mira mitigates the severe memory bottlenecks inherent to MoE inference. It enables high-throughput, high-fidelity large language model execution on widely available consumer hardware, marking a significant step toward democratizing access to state-of-the-art AI.


\bibliographystyle{IEEEtranS}
\bibliography{refs}

\end{document}